\documentclass{article}

\usepackage[preprint,nonatbib]{neurips_2022}

\usepackage[utf8]{inputenc} % allow utf-8 input
\usepackage[T1]{fontenc}    % use 8-bit T1 fonts
\usepackage{hyperref}       % hyperlinks
\usepackage{url}            % simple URL typesetting
\usepackage{booktabs}       % professional-quality tables
\usepackage{amsfonts}       % blackboard math symbols
\usepackage{nicefrac}       % compact symbols for 1/2, etc.
\usepackage{microtype}      % microtypography
\usepackage{xcolor}         % colors
\usepackage{graphicx}
\usepackage{amsmath}
\usepackage{amsthm}
\usepackage{booktabs}
\usepackage{algorithm}
\usepackage{algorithmic}
\usepackage{amssymb}
\usepackage{multirow}
\usepackage{verbatim}
\usepackage{color}
\title{Deep Reinforcement Learning with Spiking Q-learning}

\author{%
  Ding Chen \\
  Department of Computer Science and Engineering \\
  Shanghai Jiao Tong University \\
  Shanghai 200240, China \\
  \texttt{lucifer1997@sjtu.edu.cn} \\
  \And
  Peixi Peng \\
  Department of Computer Science and Technology \\
  Peking University \\
  Beijing 100871, China \\
  \texttt{pxpeng@pku.edu.cn} \\
  \AND
  Tiejun Huang \\
  Department of Computer Science and Technology \\
  Peking University \\
  \texttt{tjhuang@pku.edu.cn} \\
  \And
  Yonghong Tian \\
  Department of Computer Science and Technology \\
  Peking University \\
  Beijing 100871, China \\
  \texttt{yhtian@pku.edu.cn} \\
}

\begin{document}

\maketitle

\begin{abstract}
  With the help of special neuromorphic hardware, spiking neural networks (SNNs) are expected to realize artificial intelligence (AI) with less energy consumption. It provides a promising energy-efficient way for realistic control tasks by combining SNNs with deep reinforcement learning (RL). There are only a few existing SNN-based RL methods at present. Most of them either lack generalization ability or employ Artificial Neural Networks (ANNs) to estimate value function in training. The former needs to tune numerous hyper-parameters for each scenario, and the latter limits the application of different types of RL algorithm and ignores the large energy consumption in training. To develop a robust spike-based RL method, we draw inspiration from non-spiking interneurons found in insects and propose the deep spiking Q-network (DSQN), using the membrane voltage of non-spiking neurons as the representation of Q-value, which can directly learn robust policies from high-dimensional sensory inputs using end-to-end RL. Experiments conducted on 17 Atari games demonstrate the DSQN is effective and even outperforms  the ANN-based deep Q-network (DQN) in most games. Moreover, the experiments show superior learning stability and robustness to adversarial attacks of DSQN.
\end{abstract}

\vspace{-3mm}
\section{Introduction}
\vspace{-1mm}

Recently, guided by the brain, neuromorphic computing has emerged as one of the most promising types of computing architecture, which could realize energy-efficient AI through spike-driven communication~\cite{kuzum2013synaptic,indiveri2015memory,roy2019towards}. The research efforts of neuromorphic computing not only facilitate the emergence of large-scale neuromorphic chips~\cite{merolla2014million,davies2018loihi,furber2014spinnaker}, but also promote the development of SNNs~\cite{maass1997networks,fang2021incorporating,fang2021deep}. In this context, the field of neuromorphic computing is a close cooperation organic whole between hardware and algorithm.

An accumulating body of research studies shows that SNN can be used as energy-efficient solutions for robot control tasks with limited on-board energy resources~\cite{mahadevuni2017navigating,bing2018end,bing2018survey}. To overcome the limitations of SNN in solving high-dimensional control problems, it would be natural to combine the energy-efficiency of SNN with the optimality of deep reinforcement learning (RL), which has been proved effective in extensive control tasks~\cite{mnih2015human,silver2018general}. Since rewards are regarded as the training guidance in RL, several works~\cite{fremaux2013reinforcement,fremaux2016neuromodulated} employ reward-based learning using three-factor learning rules. However, these methods only apply to shallow SNNs and low-dimensional control tasks, or need to tune numerous hyper-parameters for each scenario~\cite{bellec2020solution}. Besides, several methods aim to apply the surrogate gradient method~\cite{lee2016training} to train SNN in RL. Since SNN usually uses the firing rate as the equivalent activation value~\cite{rueckauer2017conversion} which is a discrete value between 0 and 1, it is hard to represent the value function of RL which doesn't have a certain range in training. Therefore, several methods~\cite{patel2019improved,tan2020strategy} aim to convert the trained DQN to SNN for execution. In addition, several methods based on hybrid framework~\cite{tang2020deep,zhang2021population} utilize SNN to model policy function (i.e., a probability distribution), and resort ANN to estimate value function for auxiliary training. However, these methods limit the application of different types of RL algorithm. In addition, RL needs a large number of trials in training, and the utilization of ANN may cause more energy consumption.

To handle these challenges, it is necessary to develop a spike-based RL method where only SNN is used in both training and execution. The key issue is to design a new neural coding method to decode the spike-train into the results of value function, realizing end-to-end spike-based RL. In nature, sensory neurons receive information from the external environment and transmit it to non-spiking interneurons through action potentials~\cite{bidaye2018six}, and then change the membrane voltage of motor neurons through graded signals, so as to achieve effective locomotion. As a translational unit, non-spiking interneurons could affect the motor output according to the sensory input.

\begin{figure}[t]
\center{\includegraphics[width=0.75\linewidth]{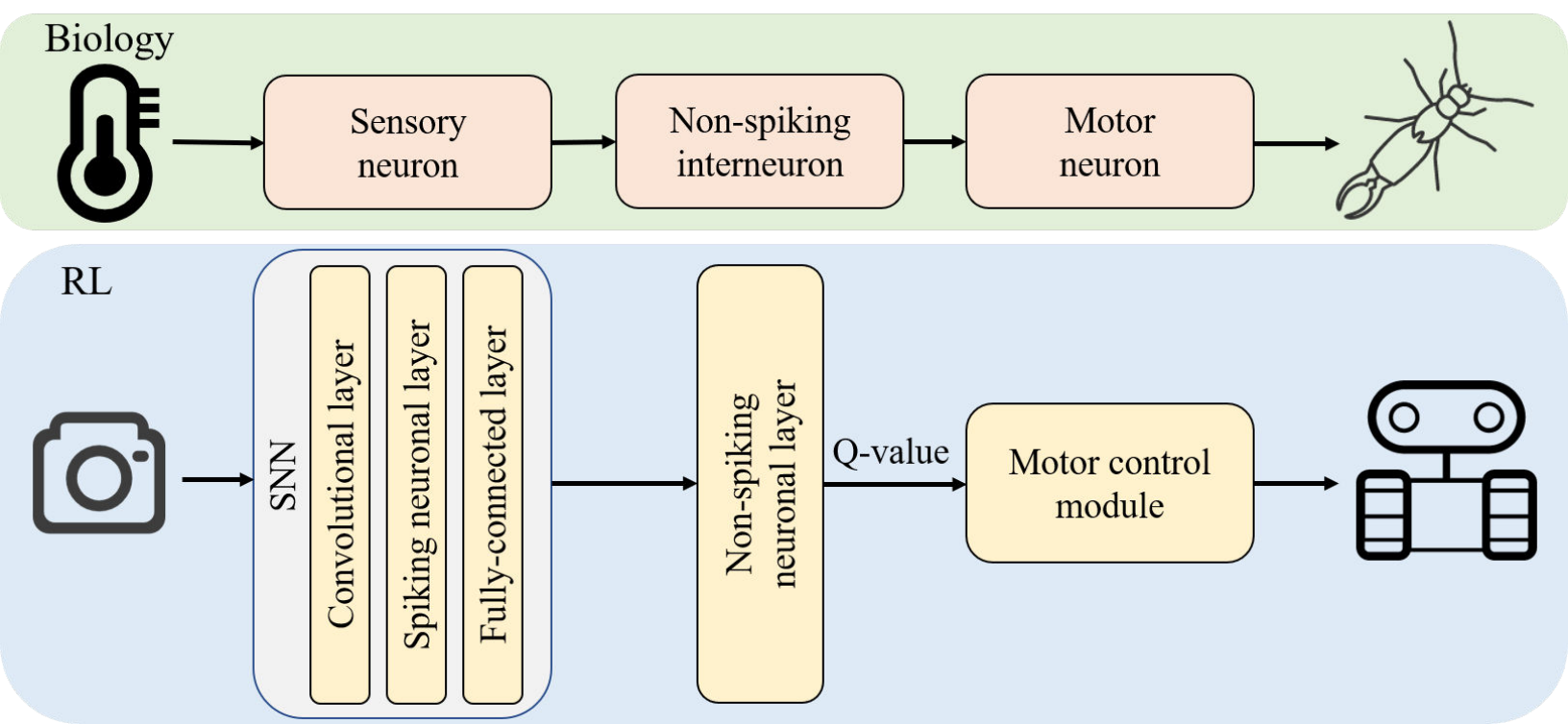}}
\caption{ \label{fig:sensorimotor} The correspondence diagram between our method and the sensory motor neuron pathway. }
\end{figure}

Inspired by biological researches on sensorimotor neuron path, we propose a novel method to train SNN with deep Q-learning~\cite{mnih2015human}, where the membrane voltage of non-spiking neurons is used to represent the Q-value (i.e., the state-action value). As shown in Figure \ref{fig:sensorimotor}, SNN receives the state from the environment and encodes them as spike-train by spiking neurons. Then, the non-spiking neurons are subsequently introduced to calculate the membrane voltage from the spike-train, which is further used to predict the Q-value of each action. Finally, the agent selects the action with the maximum Q-value. To take full advantage of the membrane voltage in all simulation time and make the learning stable, we argue that the statistics of membrane voltage rather than the last membrane voltage should be used to represent Q-value. We empirically find that the maximum membrane voltage among all the simulation time can significantly enhance spike-based RL and ensure strong robustness. The main contributions of this paper are summarized as follows:

\begin{enumerate}
    \item A novel spike-based RL method, referred to as deep spiking Q-network (DSQN), is proposed, which could represent Q-value by the membrane voltage of non-spiking neurons.
    \item The method achieves end-to-end Q-learning without any additional ANN for auxiliary training, and all calculations can be completed in neuromorphic hardware without any additional traditional hardware for auxiliary calculation, which could maximize the advantages of neuromorphic hardware to greatly reduce the overall energy consumption.
    \item The evaluation on 17 Atari games demonstrates the effectiveness of DSQN, and even DSQN outperforms ANN-based DQN on most of the games. Moreover, the experimental results show superior learning stability and robustness to adversarial attacks such as FGSM~\cite{goodfellow2014explaining}.
\end{enumerate}

\vspace{-4mm}
\section{Related Work}
\vspace{-3mm}

\paragraph{Reward-based Learning by Three-factor Learning Rules}

To bridge the gap between the time scales of behavior and neuronal action potential, modern theories of synaptic plasticity assume that the co-activation of presynaptic and postsynaptic neurons sets a flag at the synapse, called eligibility trace~\cite{sutton2018reinforcement}. Only if a third factor, signaling reward, punishment, surprise or novelty, exists while the flag is set, the synaptic weight will change. Although the theoretical framework of three-factor learning rules has been developed in the past few decades, experimental evidence supporting eligibility trace has only been collected in the past few years~\cite{gerstner2018eligibility}. Through the derivation of the RL rule for continuous time, the existing approaches have been able to solve the standard control tasks~\cite{fremaux2013reinforcement} and robot control tasks~\cite{mahadevuni2017navigating}. However, these methods are only suitable for shallow SNNs and low-dimensional control tasks. To solve these problems, Bellec {\it et al.}~\cite{bellec2020solution} propose a learning method for recurrently connected networks of spiking neurons, which is called e-prop. Despite the agent learned by reward-based e-prop successfully wins Atari games, the need that numerous hyper-parameters should be adjusted between different tasks limits the application of this method.

\paragraph{ANN to SNN Conversion for RL}

By matching the firing rate of spiking neurons with the graded activation of analog neurons, trained ANNs can be converted into corresponding SNNs with few accuracy loss~\cite{rueckauer2017conversion}. For the SNNs converted from the ANNs trained by DQN algorithm, the firing rate of spiking neurons in the output layer is proportional to the Q-value of the corresponding action, which makes it possible to select actions according to the relative size of the Q-value~\cite{patel2019improved,tan2020strategy}. But there is a trade-off between the accuracy and the efficiency, which tells us that longer inference latency is needed to improve accuracy. As far as we know, for RL tasks, the converted SNNs cannot achieve better results than ANNs.

\paragraph{Co-learning of Hybrid Framework}

Tang {\it et al.}~\cite{tang2020reinforcement} first propose the hybrid framework, composed of a spiking actor network (SAN) and a deep critic network. Through the co-learning of the two networks, these methods ~\cite{tang2020deep,zhang2021population} avoid the problem of value estimation using SNNs. Hence, these methods only work for actor-critic structure, and the energy consumption in the training process is much higher than the pure SNN methods~\cite{kim2021chip}.

\paragraph{RL methods using Spike-based BP}

Following the surrogate gradient method~\cite{lee2016training}, the spike-based backpropagation (BP) algorithm has quickly become the mainstream solution for training multi-layer SNNs~\cite{fang2021incorporating,fang2021deep}. Recently, Liu {\it et al.}~\cite{liu2021human} propose a direct spiking
learning algorithm for the deep spiking Q-network, using a fully-connected layer to decode the firing rate into Q-value. Different from them, we use the membrane voltage of non-spiking neurons to represent the Q-value, so that our method can be directly applied to neuromorphic hardware (see the Section \ref{sec:analysis} for a detailed description). As shown in temporal coding benchmarks~\cite{cramer2020heidelberg} and open-source frameworks~\cite{SpikingJelly,norse2021}, the membrane voltage of non-spiking neurons is feasible to represent a continuous value in a spike-based BP method. However, how to use it to effectively train SNN with Q-learning has not been systematically studied and remains unsolved, which is the goal of this paper.

\vspace{-4mm}
\section{Method}
\vspace{-3mm}

In this section, we present the deep spiking Q-network (DSQN) in detail. Firstly, we introduce the preliminary of RL, the spiking neural model and its discrete dynamics subsequently. Then, we propose the non-spiking neurons for DSQN and analyze the neural coding method. Finally, we present the learning algorithm of spike-based RL and derive the detailed learning rule.

\vspace{-3mm}
\subsection{Deep Q-learning}
\vspace{-1mm}

The goal of RL algorithm is to train a strategy to maximize the expected cumulative reward in a Markov decision process (MDP). In a RL task, the agent interacts with the environment through a series of observations ($s$), actions ($a$) and rewards ($r$). The Q-value function $Q(s,a)$ is to estimate the expected cumulative rewards of performing action $a$ at $s$. We refer to a neural network function approximator with weights $\theta$ as the Q-network, which can be learned by minimizing the mean-squared error in the Bellman equation in training. Therefore, the loss function is defined as follows:
\begin{small} 
\begin{align}
\label{eq:loss_func}
    L\left(\theta\right)=&\mathbb{E}_{s,a,r}\left[\left(\mathbb{E}_{s^\prime}\left[y\middle| s,a\right]-Q\left(s,a;\theta\right)\right)^2\right]\nonumber\\
    =&\mathbb{E}_{s,a,r,s^\prime}\left[\left(y-Q\left(s,a;\theta\right)\right)^2\right]+\mathbb{E}_{s,a,r}\left[\mathbb{V}_{s^\prime}\left[y\right]\right],
\end{align}
\end{small}
where $y=r+\gamma\max\limits_{a^\prime}Q(s^\prime,a^\prime;\theta^{-})$. $s^\prime$ and $a^\prime$ represent the observation and the action at the next time-step, $\gamma$ is a discount factor and $\theta^{-}$ are the weights of the target network. $\mathbb{E}$ and $\mathbb{V}$ represent the expectation and the variance. More details of Eq. (\ref{eq:loss_func}) refer to DQN~\cite{mnih2015human}. Differentiating the loss function with respect to the weights, we obtain the following gradient:
\begin{small}
\begin{align}
\label{eq:loss_grad}
    \nabla_\theta L\left(\theta\right)=\mathbb{E}_{s,a,r,s^\prime}\left[\left(y-Q\left(s,a;\theta\right)\right)\nabla_\theta Q\left(s,a;\theta\right)\right].
\end{align}
\end{small}

\begin{figure}[t]
\center{\includegraphics[width=0.8\linewidth]{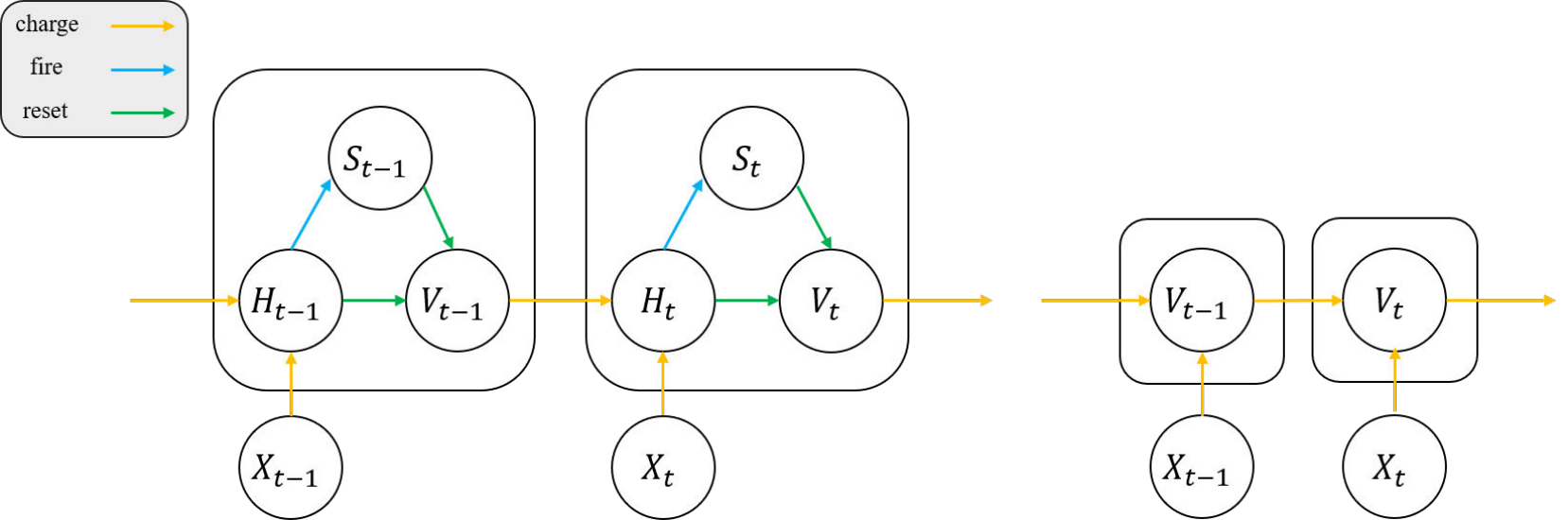}} 
\caption{ \label{fig:neural_model} The general neural model (Left) Spiking neural model. (Right) Non-spiking neural model. }
\end{figure}

\vspace{-3mm}
\subsection{Spiking Neural Model}
\vspace{-1mm}

The basic computing unit of SNN is the spiking neuron. The dynamics of all kinds of spiking neurons can be described as:
\begin{small}
\begin{align}
\label{eq:charging}
    H_t &=f\left(V_{t-1},X_t\right),\\
\label{eq:firing}
    S_t &=\mathrm{\Theta}\left(H_t-V_{th}\right),\\
\label{eq:resetting}
    V_t &=H_t\left(1\ -\ S_t\right)+V_{reset}S_t,
\end{align}
\end{small}
where $H_t$ and $V_t$ denote the membrane voltage after neural dynamics and the trigger of a spike at time-step $t$, respectively. $X_t$ denotes the external input, and $S_t$ means the output spike at time-step $t$, which equals 1 if there is a spike and 0 otherwise. $V_{th}$ denotes the threshold voltage and $V_{reset}$ denotes the membrane rest voltage. As shown in Figure \ref{fig:neural_model}, Eq. (\ref{eq:charging}) - (\ref{eq:resetting}) establish a general mathematical model to describe the discrete dynamics of spiking neurons, which include charging, firing and resetting. Specifically, Eq. (\ref{eq:charging}) describes the subthreshold dynamics, which vary with the type of neuron models. Here we consider the Leaky Integrate-and-Fire (LIF) model~\cite{gerstner2014neuronal}, which is one of the most common spiking neuron models. The function $f(\cdot)$ of the LIF neuron is defined as:
\begin{small}
\begin{equation}
\label{eq:LIF}
    f\left(V_{t-1},X_t\right)=V_{t-1}+\frac{1}{\tau}\left(-\left(V_{t-1}-V_{reset}\right)+X_t\right),
\end{equation}
\end{small}
where $\tau$ is the membrane time constant. The spike generative function $\Theta(x)$ is the Heaviside step function, which is defined by $\Theta(x)=1$ for $x\geq 0$ and $\Theta(x)=0$ for $x< 0$. Note that $ V_0=V_{reset}$.

\vspace{-3mm}
\subsection{Non-spiking Neural Model}
\vspace{-1mm}

Non-spiking neurons can be regarded as a special case of spiking neurons. If we set the threshold voltage $V_{th}$ of spiking neurons to infinity, the dynamics of neurons will always be under the threshold, which is so-called non-spiking neurons. Since non-spiking neurons do not have the dynamics of firing and resetting, we could simplify the neural model to the following equation (see Figure \ref{fig:neural_model}):
\begin{small}
\begin{equation}
\label{eq:nonspiking}
    V_t=f\left(V_{t-1},X_t\right).
\end{equation}
\end{small}

Here we consider the non-spiking LIF model, which can also be called Leaky Integrate (LI) model. The dynamics of LI neurons is described as the following equation:
\begin{small}
\begin{equation}
\label{eq:LI}
    V_t=V_{t-1}+\frac{1}{\tau}\left(-\left(V_{t-1}-V_{reset}\right)+X_t\right).
\end{equation}
\end{small}

\vspace{-3mm}
\subsection{Neural Coding}
\vspace{-1mm}

According to the definition of SNN, the output is a spike-train, but the results of value estimation used in RL are continuous values. To bridge the difference between these two data forms, we need a spike decoder to complete the data conversion, that is non-spiking neurons.
In the sensorimotor neuron path, the membrane voltage of non-spiking interneurons determines the input current of motor neurons. Therefore, for different non-spiking neurons, the greater the membrane voltage, the greater the probability of taking the corresponding action, which reminds us of the Q-value function in RL.

In the whole simulation time $T$, the non-spiking neurons take the spike-train as the input sequence, and then the membrane voltage $V_t$ at each time-step $t$ can be obtained. To represent the output of value function, we need to choose an optimal statistic according to the membrane voltage at all times.

To meet the needs of the algorithm, we finally design three statistics as candidates: 
\begin{itemize}
\item \textbf{Last membrane voltage (last-mem)}: It is a natural idea to use the last membrane voltage after the simulation time as the characterization of the Q-value ($Q=V_T$) to make full use of all simulation.
\item \textbf{Maximum membrane voltage (max-mem)}: By recording the membrane voltage of non-spiking neurons at each time-step in the whole simulation time, we can get the maximum membrane voltage ($Q=\max _{1\le t\le T}{V_t}$).
\item \textbf{Mean membrane voltage (mean-mem)}: Similar to the maximum membrane voltage, we can obtain the mean value by collecting the membrane voltage data, which is also a meaningful statistic ($Q=\frac{1}{T}\sum_{t=1}^{T}V_t$).
\end{itemize}

\begin{figure}[t]
\center{\includegraphics[width=0.8\linewidth]{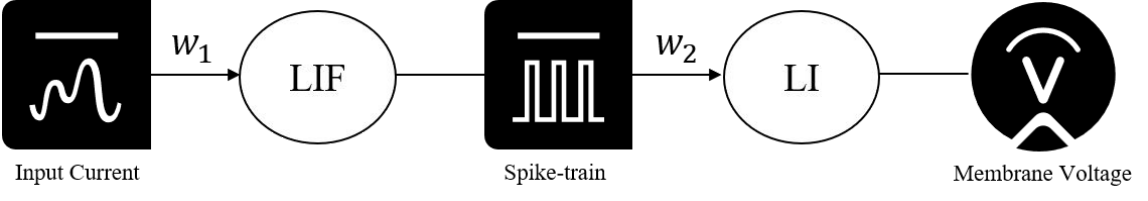}} 
\caption{ \label{fig:simple_case} The architecture of the simple network for the case study on different statistics. }
\end{figure}

\begin{figure}[t]
\center{\includegraphics[width=0.7\linewidth]{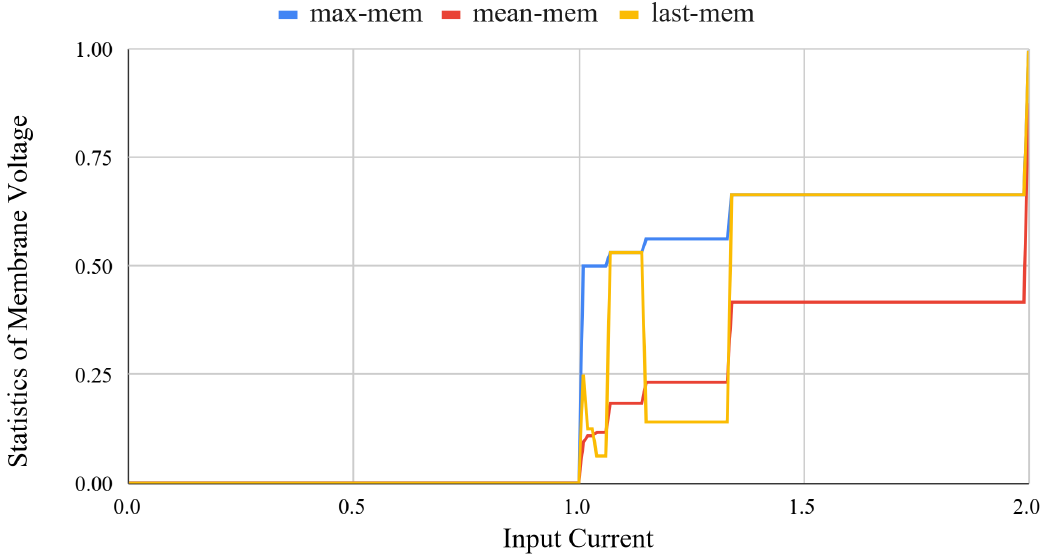}} 
\caption{ \label{fig:function} The functional relationship between input current and statistics of membrane voltage. }
\end{figure}

To have an intuitive understanding of the neuronal dynamics of non-spiking neurons decoded with different statistics, we conduct a case study to choose the decoder used in our methods. As illustrated in Figure \ref{fig:simple_case}, we consider a simple network as an example, which consists of a LIF neuron and a LI neuron. The LIF neuron receives the weighted input $X(t)=w_1I(t)$, and then transmits the weighted spike signal $w_2S(t)$ to the LI neuron. In this case, the input $I(t)$ is a constant, and the simulation time $T$ is set to 8. Furthermore, the membrane rest voltage $V_{reset}$ is set to 0, and the membrane time constant $\tau$ is set to 2.0. For simplification, the weights $w_1$ and $w_2$ are set to 1.0. In Figure \ref{fig:function}, we plot the functional relationship between the input current and different statistics of membrane voltage. 

As we can see, with the slight increase of input current, sometimes the firing time of LIF neurons will be advanced without generating more spikes, which leads to further attenuation of the membrane voltage caused by the spikes from the previous layer. So the membrane voltage of LI neurons at the last moment may be much lower, although the input current increases. From the perspective of optimization, the last membrane voltage may cause the network output nonperiodic oscillatory, make the learning unstable and hard to converge. Hence, we remove the last-mem from the candidates. 
For max-mem and mean-mem, we empirically evaluate them in experiments respectively in Section \ref{sec:analysis}. We find both of them are effective and the max-mem performs better. Hence, we choose max-mem as the decoding method in the following sections.

\begin{figure}[t]
\center{\includegraphics[width=0.8\linewidth]{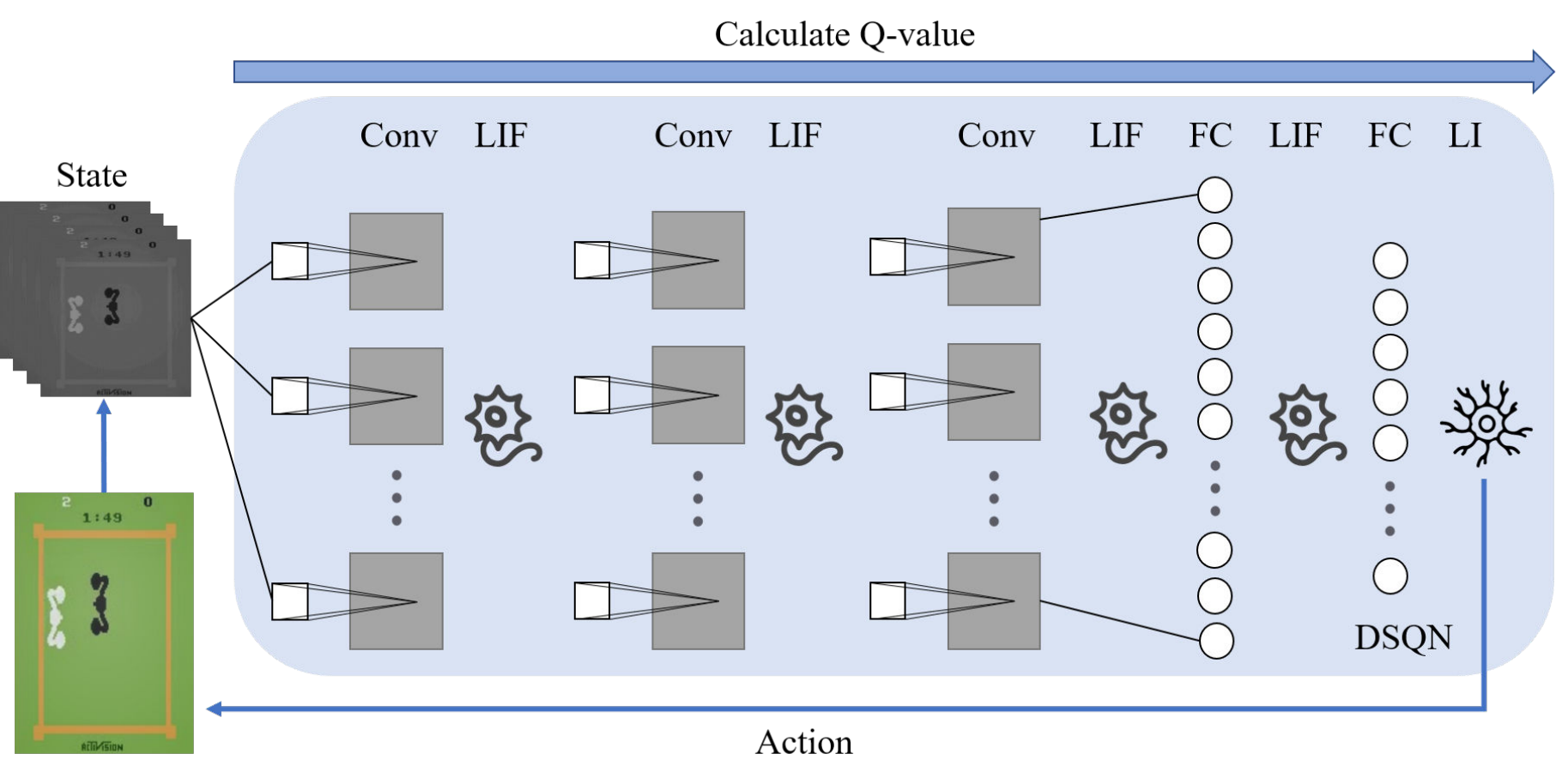}} 
\caption{ \label{fig:dsqn} The architecture of deep spiking Q-network. }
\end{figure}

\vspace{-3mm}
\subsection{Deep Spiking Q-network}
\vspace{-2mm}

The typical architecture of DSQN is shown in Figure \ref{fig:dsqn}, which consists of synaptic layers and neuronal layers. The synaptic layers include convolutional layers and fully-connected (FC) layers, each of which is followed by a neuronal layer. Except that the LI layer is used after the last FC layer, the other synaptic layers are followed by LIF layers. In DSQN, all bias parameters are omitted and all weight parameters are shared at all simulation time-steps.

Note that we treat the synaptic layer and its subsequent neuronal layer as one layer in formula derivation. The former plays a  similar role to dendrites in neuronal cells, and the latter works like the bodies and axons. Therefore, the first layer acts as a learnable spike encoder to convert the static state into the corresponding spike-train. During the task, the agent selects the action according to the Q-value calculated by the network in the simulation time, and then obtains the new environmental state to make the next decision. 

\vspace{-3mm}
\subsection{Training Framework}
\vspace{-2mm}

Here we derive the detailed backpropagation training algorithm for DSQN. According to Eq. (\ref{eq:loss_grad}), we just need to derive the gradient of Q-value with respect to the weights using the network formula. Suppose that $\theta^i$ represents the weight parameter of the {\it i}-th layer in the network ($1\le i\le l$). For {\it i}-th layer, $I_t^i$ and $S_t^i$ represent the input and output signal at time-step $t$, which ensure that $I_t^i=S_t^{i-1}$ if $i>1$. The external input of neuronal model is $X_t^i=\theta^{i}I_t^i$. $\Theta^\prime(x)$ is defined by the surrogate function $\Theta^\prime(x)=\sigma^\prime(x)$. Taking the DSQN algorithm (max-mem) as an example, we can calculate the gradients recursively:
\begin{small}
\begin{equation}
\label{eq:T_max}
    t^\prime=\mathop{\arg\max}\limits_{1\le t\le T}{V_t^l},
\end{equation}
\end{small}
\begin{small}
\begin{equation}
\label{eq:V_grad}
    \frac{\delta V_t^l}{\delta\theta^l}=\left\{\begin{matrix}\frac{\delta V_t^l}{\delta V_{t-1}^l}\frac{\delta V_{t-1}^l}{\delta\theta^l}+\frac{\delta V_t^l}{\delta X_t^l}\frac{\delta X_t^l}{\delta\theta^l}&if\ t>1\\\frac{\delta V_t^l}{\delta X_t^l}\frac{\delta X_t^l}{\delta\theta^l}&if\ t=1\\\end{matrix}\right..
\end{equation}
\end{small}

According to Eq. (\ref{eq:charging}) - (\ref{eq:resetting}), we can get:
\begin{small}
\begin{equation}
\label{eq:Q_H}
    \frac{\delta Q}{\delta H_t^i}=\left\{\begin{matrix}\frac{\delta Q}{\delta H_{t+1}^i}\frac{\delta H_{t+1}^i}{\delta H_t^i}+\frac{\delta V_{t^\prime}^l}{\delta S_t^{l-1}}\frac{\delta S_t^{l-1}}{\delta H_t^i}&if\ t<t^\prime\\\frac{\delta V_t^l}{\delta S_t^{l-1}}\frac{\delta S_t^{l-1}}{\delta H_t^i}&if\ t=t^\prime\\0&if\ t>t^\prime\\\end{matrix}\right.,
\end{equation}
\end{small}
\begin{small}
\begin{equation}
\label{eq:H_H}
    \frac{\delta H_{t+1}^i}{\delta H_t^i}=\frac{\delta H_{t+1}^i}{\delta V_t^i}\frac{\delta V_t^i}{\delta H_t^i}=\left(1-\frac{1}{\tau}\right)\frac{\delta V_t^i}{\delta H_t^i},
\end{equation}
\end{small}
\begin{small}
\begin{equation}
\label{eq:V_S}
    \frac{\delta V_{t^\prime}^l}{\delta S_t^{l-1}}=\frac{\delta V_{t^\prime}^l}{\delta V_t^l}\frac{\delta V_t^l}{\delta X_t^l}\frac{\delta X_t^l}{\delta S_t^{l-1}}=\left(1-\frac{1}{\tau}\right)^{t^\prime-t}\frac{\theta^l}{\tau},
\end{equation}
\end{small}
\begin{small}
\begin{equation}
\label{eq:S_H}
    \frac{\delta S_t^j}{\delta H_t^i}=\left\{\begin{matrix}\frac{\delta S_t^j}{\delta H_t^j}\frac{\delta H_t^j}{\delta X_t^j}\frac{\delta X_t^j}{\delta S_t^{j-1}}\frac{\delta S_t^{j-1}}{\delta H_t^i}&if\ j>i\\\sigma^\prime(H_t^j-V_{th})&if\ j=i\\\end{matrix}\right.,
\end{equation}
\end{small}
\begin{equation}
\label{eq:V_H}
    \frac{\delta V_t^i}{\delta H_t^i}=1-S_t^i+\left(V_{reset}-H_t^i\right)\frac{\delta S_t^i}{\delta H_t^i},
\end{equation}
\begin{small}
\begin{equation}
\label{eq:H_theta}
    \frac{\delta H_t^i}{\delta\theta^i}=\left\{\begin{matrix}\frac{\delta H_t^i}{\delta V_{t-1}^i}\frac{\delta V_{t-1}^i}{\delta H_{t-1}^i}\frac{\delta H_{t-1}^i}{\delta\theta^i}+\frac{\delta H_t^i}{\delta X_t^i}\frac{\delta X_t^i}{\delta\theta^i}&if\ t>1\\\frac{\delta H_t^i}{\delta X_t^i}\frac{\delta X_t^i}{\delta\theta^i}&if\ t=1\\\end{matrix}\right..
\end{equation}
\end{small}

Finally, we can get the gradients of the weight parameters:
\begin{small}
\begin{equation}
\label{eq:Q_theta}
    \frac{\delta Q}{\delta\theta^i}=\left\{\begin{matrix}\sum_{t=1}^{T}{\frac{\delta Q}{\delta H_t^i}\frac{\delta H_t^i}{\delta\theta^i}}&i<l\\\frac{\delta V_{t^\prime}^i}{\delta\theta^i}&i=l\\\end{matrix}\right..
\end{equation}
\end{small}
The gradient of mean-mem based DSQN could be derived in a similar way.

\vspace{-3mm}
\section{Experiments}
\vspace{-2mm}

In this section, we firstly compare the different decoders to choose, and evaluate the performance of DSQN on 17 Atari 2600 games from the arcade learning environment (ALE)~\cite{bellemare2013arcade}. Furthermore, we theoretically compare the energy consumption of each layer in DSQN with DQN.  Finally, we experimentally evaluate the robustness of DSQN when dealing with adversarial attacks. Our implementation of DSQN is based on the open-source framework of SNN~\cite{SpikingJelly}.

\vspace{-3mm}
\subsection{Experimental Settings}
\vspace{-2mm}

For Atari games, the ANN used in DQN~\cite{mnih2015human} contains 3 convolutional layers and 2 FC layers (i.e., Input-32C8S4-ReLU-64C4S2-ReLU-64C3S1-ReLU-Flatten-512-ReLU-$N_{A}$, where $N_{A}$ is the number of actions used in the task). For a fair comparison, DSQN employs a similar network structure for fair comparison (i.e., Input-32C8S4-LIF-64C4S2-LIF-64C3S1-LIF-Flatten-512-LIF-$N_{A}$-LI). The arctangent function is used as the surrogate function. The gradient is defined by
\begin{small}
\begin{equation}
\label{eq:surrogate_grad}
    \sigma^\prime(x)=\frac{1}{1+(\pi x)^{2}},
\end{equation}
\end{small}
where the primitive function is
\begin{small}
\begin{equation}
\label{eq:surrogate_func}
    \sigma(x)=\frac{1}{\pi}\arctan{\pi x} + \frac{1}{2}.
\end{equation}
\end{small}
Due to the limitation of resources, we choose 17 top-performing Atari games selected by~\cite{tan2020strategy} to test the method. The network architecture and hyper-parameters keep identical across all 17 games. The hyper-parameters are same with~\cite{mnih2015human}, except that we use the Adam optimizer and train for a total of 20 million frames. This change is to improve the training efficiency of the tasks on the premise of ensuring high performance. Detailed tables of hyper-parameters are listed in the appendix.

\vspace{-3mm}
\subsection{Performance}
\vspace{-1mm}

To obtain the optimal performance of each method, we evaluate the model every 100,000 frames during the training process, with a total of 30 episodes of each test. The training process is repeated 3 times with different random seeds. During the evaluation, the agent starts with a random number (up to 30 times) of no-op actions in each episode, and the behavior policy is $\epsilon$-greedy with $\epsilon$ fixed at 0.05. Due to space constraints, here we only report the optimal performance of different models and several typical learning curves. The complete learning curves are shown in the appendix as well as the raw and normalized scores.

\begin{table}[t]
\caption{The performance of Atari games obtained by DSQN with different decoders.}
\label{tab:decoders}
\centering
\begin{tabular}{lrrr}
\toprule
Game  & max-mem & mean-mem & fr-decoder~\cite{cramer2020heidelberg} \\
\midrule
Atlantis    & 2515926.7 $\pm$ 73782.9 & 2319833.3 $\pm$ 446159.2 & \textbf{2602666.7 $\pm$ 155432.4} \\
Beam Rider  & 5327.6 $\pm$ 178.0 & 5098.3 $\pm$ 222.1 & \textbf{5747.6 $\pm$ 267.0} \\
Boxing      & \textbf{82.7 $\pm$ 8.7} & 65.1 $\pm$ 9.5 & 71.4 $\pm$ 8.8 \\
Breakout    & \textbf{368.1 $\pm$ 9.8} & 352.8 $\pm$ 12.3 & 359.3 $\pm$ 5.5 \\
\bottomrule
\end{tabular}
\end{table}

\begin{table}[t]
\caption{The performance of Atari games obtained by DQN, ANN-SNN and DSQN.}
\label{tab:perf}
\centering
\begin{tabular}{lrrr}
\toprule
Game  & DQN Score ($\pm$ std) & ANN-SNN Score ($\pm$ std) & DSQN Score ($\pm$ std) \\
\midrule
Atlantis & \textbf{2771856.7 $\pm$ 177411.9} & 1590565.5 $\pm$ 1025814.5 & 2515926.7 $\pm$ 73782.9 \\
Beam Rider & \textbf{5841.7 $\pm$ 982.7} & 4905.9 $\pm$ 840.6 & 5327.6 $\pm$ 178.0 \\
Boxing & 21.2 $\pm$ 6.7 & 17.8 $\pm$ 5.0 & \textbf{82.7 $\pm$ 8.7} \\
Breakout & 261.4 $\pm$ 11.6 & 203.9 $\pm$ 23.9 & \textbf{368.1 $\pm$ 9.8} \\
Crazy Climber & 85380.0 $\pm$ 15448.3 & 73753.0 $\pm$ 10951.8 & \textbf{95164.4 $\pm$ 1232.9} \\
Gopher & 3204.5 $\pm$ 394.4 & 2334.2 $\pm$ 852.9 & \textbf{4233.1 $\pm$ 176.5} \\
Jamesbond & \textbf{518.3 $\pm$ 165.4} & 327.5 $\pm$ 170.3 & 469.4 $\pm$ 82.7 \\
Kangaroo & 4793.3 $\pm$ 2649.9 & 4033.3 $\pm$ 2280.4 & \textbf{5824.4 $\pm$ 540.8} \\
Krull & \textbf{34698.3 $\pm$ 37315.8} & 27597.1 $\pm$ 31259.4 & 6991.1 $\pm$ 107.0 \\
Name This Game & 4991.8 $\pm$ 302.4 & 4521.2 $\pm$ 143.2 & \textbf{6981.0 $\pm$ 192.8} \\
Pong & -20.5 $\pm$ 0.1 & -20.7 $\pm$ 0.02 & \textbf{19.5 $\pm$ 0.4} \\
Road Runner & 2600.0 $\pm$ 604.7 & 2271.1 $\pm$ 2027.0 & \textbf{27725.6 $\pm$ 3954.0} \\
Space Invaders & 1040.2 $\pm$ 17.4 & 867.2 $\pm$ 111.8 & \textbf{1202.9 $\pm$ 61.2} \\
Star Gunner & 1133.3 $\pm$ 115.9 & 1081.1 $\pm$ 54.0 & \textbf{1657.8 $\pm$ 102.0} \\
Tennis & -1.0 $\pm$ 0.0 & -1.0 $\pm$ 0.0 & -1.0 $\pm$ 0.0 \\
Tutankham & 156.5 $\pm$ 78.6 & 133.3 $\pm$ 87.5 & \textbf{266.5 $\pm$ 12.2} \\
Video Pinball & 310719.8 $\pm$ 132048.9 & 12533.7 $\pm$ 14853.7 & \textbf{408032.6 $\pm$ 41687.8} \\
\bottomrule
\end{tabular}
\end{table} 

\vspace{-3mm}
\subsubsection{Analysis of Decoders}
\label{sec:analysis}
\vspace{-1mm}

We compare the performance of DSQN with different decoders on four games  selected in alphabetical order (Table \ref{tab:decoders}). In all of these games, DSQN (max-mem) achieves better performance than DSQN (mean-mem). Hence we use the maximum membrane voltage in the subsequent experiments. In terms of performance, DSQN (fr-decoder)~\cite{cramer2020heidelberg} is roughly equivalent to DSQN (max-mem). However, since the neuromorphic chips only support spike-driven computations, the firing rate needs to be calculated with the help of other traditional hardware to estimate the Q-value in DSQN (fr-decoder). For our method, we only need to utilize the existing components of neuromorphic chip, i.e., the amount of charge stored in the capacitor is used to represent the Q-value~\cite{kim2021chip}.
 
\begin{table*}[t]
\caption{The mean scores of DQN and DSQN when dealing with adversarial attacks. Since the decay rate cannot be used for comparison when the reward is negative, some values are omitted.}
\label{tab:FGSM}
\centering
\begin{tabular}{lrrrrrr}
\toprule
\multicolumn{1}{c}{} & \multicolumn{3}{c}{DQN} & \multicolumn{3}{c}{DSQN} \\
\multicolumn{1}{c}{\multirow{-2}{*}{Game}} & before & after & decay rate & before & after & decay rate \\
\midrule
Atlantis & 2945880.0 & 22543.3 & 99.23\% & 2481620.0 & 34666.7 & \textbf{98.60\%} \\
Beam Rider & 5211.5 & 2055.0 & 60.57\% & 5188.9 & 4198.0 & \textbf{19.10\%} \\
Boxing & 28.3 & 15.4 & 45.70\% & 84.4 & 77.5 & \textbf{8.14\%} \\
Breakout & 249.9 & 1.2 & 99.53\% & 360.8 & 19.3 & \textbf{94.65\%} \\
Crazy Climber & 85396.7 & 19390.0 & 77.29\% & 93753.3 & 82933.3 & \textbf{11.54\%} \\
Gopher & 2788.7 & 580.7 & 79.18\% & 4154.0 & 1780.5 & \textbf{57.14\%} \\
Jamesbond & 406.7 & 201.7 & 50.41\% & 463.3 & 350.0 & \textbf{24.46\%} \\
Kangaroo & 6493.3 & 1293.3 & 80.08\% & 6140.0 & 1754.5 & \textbf{71.43\%} \\
Krull & 24042.9 & 21204.0 & 11.81\% & 6899.0 & 6528.3 & \textbf{5.37\%} \\
Name This Game & 4647.0 & 1796.3 & 61.34\% & 7082.7 & 6367.7 & \textbf{10.10\%} \\
Pong & -20.5 & -20.7 & - & 19.1 & 4.8 & 74.87\% \\
Road Runner & 2530.0 & 2023.3 & 20.03\% & 23206.7 & 20666.7 & \textbf{10.94\%} \\
Space Invaders & 1059.7 & 540.2 & 49.02\% & 1132.3 & 915.0 & \textbf{19.19\%} \\
Star Gunner & 1116.7 & 1013.3 & 9.25\% & 1716.7 & 1713.9 & \textbf{0.16\%} \\
Tennis & -1.0 & -1.0 & 0\% & -1.0 & -1.0 & 0\% \\
Tutankham & 98.4 & 15.6 & 84.17\% & 276.0 & 116.0 & \textbf{57.97\%} \\
Video Pinball & 276040.2 & 43437.2 & 84.26\% & 441615.2 & 441592.7 & \textbf{0.01\%} \\
\bottomrule
\end{tabular}
\end{table*}

\begin{figure*}[t]
\center{\includegraphics[width=0.95\linewidth]{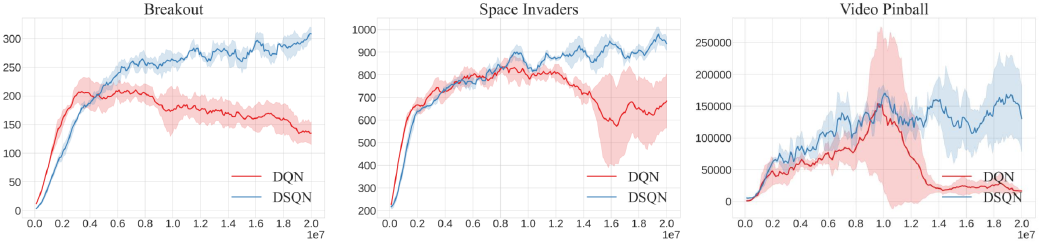}} 
\caption{ \label{fig:curves} Learning curves of DQN and DSQN. Each curve is smoothed with a moving average of 10 to improve readability. }
\end{figure*}

\vspace{-3mm}
\subsubsection{The Performance of DSQNs on Atari Games}
\vspace{-1mm}

The existing co-learning methods of hybrid framework aim to use vector state and continuous action space (simulated on MuJoCo~\cite{todorov2012mujoco}) to solve continuous control tasks, which is difficult to migrate directly to tasks with image state and discrete action space (i.e. Atari game). Therefore, the DSQN is compared with DQN and Converted SNN (ANN-SNN)~\cite{tan2020strategy}, where DQN and ANN-SNN are re-run under the same experimental setting for a fair comparison. As shown in Table \ref{tab:perf}, DSQN achieves better or comparable performance with DQN on most games. Moreover, the learning of DSQN is more stable and could handle over-estimation well on most games, as shown in Figure \ref{fig:curves} (see the appendix for the learning curves of all Atari games). It indicates DSQN could control an agent effectively, and provides a promising way to the energy-efficient control tasks. The simulation time $T$ is set to 500 (ANN-SNN) and 8 (DSQN), respectively. Therefore, DSQN outperforms ANN-SNN on most games using fewer simulation time-steps. The reason is that ANN-SNN is limited by the trained ANN, while DSQN is trained directly from the environment.

\vspace{-2mm}
\subsection{Computational Cost}
\vspace{-1mm}

To estimate the energy efficiency of DSQN and compare it with DQN, we calculate the amount of calculation by accumulation (AC) and multiply-and-accumulate (MAC) operations. The main energy consumption of ANNs comes from the MAC between neurons. In contrast, for SNNs, a transmitted spike requires only an AC at the target neuron, adding the weight to the membrane voltage. Furthermore, the spiking neurons update the internal state at every time-step at the cost of several MACs depending on their complexity~\cite{roy2019towards}. With the deepening of network structure, the relative energy ratio gradually approaches a fixed value, which could be calculated as follows:
\begin{small}
\begin{equation}
\label{eq:energy_ratio}
    \frac{E(SNN)}{E(ANN)}\approx T*fr*\frac{E(AC)}{E(MAC)},
\end{equation}
\end{small}
where $T$ and $fr$ represent the simulation time and the average firing rate. For a 45 nm complementary metal-oxide-semiconductor, $E(MAC) = 31 E(AC)$~\cite{horowitz20141}. In our experiment, $T$ is set to 8 and $fr$ is around 0.1-0.2. Compared with the ANNs, the use of SNNs on neuromorphic hardware can reduce 94.84\%-97.43\% of the energy consumption per inference. The detailed analysis of theoretical energy computation is described in the appendix.

\vspace{-2mm}
\subsection{Robustness to White-box Attacks}
\label{sec:attack}
\vspace{-1mm}

Previous study~\cite{patel2019improved} shows that SNN could improve robustness to occlusion in the input image for RL tasks. Moreover, quite a few articles~\cite{sharmin2019comprehensive,sharmin2020inherent,kim2021visual} have proved the robustness of SNNs in other fields. To verify the robustness of DSQN, DSQN and DQN are evaluated under the white-box attacks respectively. Due to the large number of frames used in RL tasks, the Fast Gradient Sign Method (FGSM)~\cite{goodfellow2014explaining} is used to generate adversarial examples efficiently. Given a Q-network with parameters $\theta$ and loss $L(\theta,x,y)$, where $x$ is the input state and $y$ is a softmax of the computed Q-values, using FGSM results in a perturbation of
\begin{small}
\begin{equation}
\label{eq:FGSM}
    \eta = \epsilon\,sign\left(\nabla_x L(\theta,x,y)\right).
\end{equation}
\end{small}

We change the strategy of the Q-network to deterministic, and the goal of the white-box attack is to make agents change the action selection by generating adversarial perturbations iteratively. The iteration is set to 50 at most.

% To save computing resources and improve efficiency, we do not guarantee the success rate of each attack, and set the upper limit of the iteration to 50.

Table \ref{tab:FGSM} shows the performances of a trained DQN and DSQN dealing with adversarial attacks respectively, where the decay rate represents the percentage of decline in the score of an agent after encountering adversarial attacks. We use the identical seed in all experiments (seed=12). Experimental results demonstrate the superior robustness of the proposed DSQN.

\vspace{-2mm}
\section{Conclusion}
\vspace{-1mm}

This paper presents a novel deep spiking Q-network to train SNN in RL task, by directly representing the Q-value with the maximum membrane voltage of non-spiking neurons. The method is evaluated on 17 Atari games, and outperforms ANN-based DQN in most scenarios. In addition, the experimental results indicate the robustness of the method under white-box attacks. The theoretical proof of performance will be explored in the future. By running DSQN method on neuromorphic hardware to process the output of neuromorphic sensors, it will have a profound social impact on the field of autonomous control and robotics. We hope that this work can pave the way for the SNN-based RL. 

\bibliographystyle{plain}
\bibliography{main}

%%%%%%%%%%%%%%%%%%%%%%%%%%%%%%%%%%%%%%%%%%%%%%%%%%%%%%%%%%%%

\appendix

\section{Appendix}

\subsection{Reproducibility}

All of the source codes will be available in the final camera-ready version for the sake of anonymity. The results are obtained by running DQNs and DSQNs with 3 different random seeds with the same hyper-parameters. To maximize reproducibility, we use identical seeds in all experiments (seed=1,12,123). The hyperparameters of DSQN are illustrated in Table \ref{tab:hyper}. Note that the models used for robustness test are trained with the same seed (seed=12). 

\begin{table}[H]
\caption{The hyperparameters of DSQN.}
\label{tab:hyper}
\centering
\begin{tabular}{lr}
\toprule
Parameter  & Value \\
\midrule
minibatch size      & 32 \\
replay start size   & 50000 \\
replay memory size  & 1000000 \\
target network update frequency & 10000 \\
Adam learning rate  & 0.00025 \\
Adam $\epsilon$     & 1e-8 \\
initial exploration & 1.0 \\
final exploration   & 0.1 \\
final exploration frame & 1000000 \\
no-op max   & 30 \\
\bottomrule
\end{tabular}
\end{table}

\subsection{Complete Learning Curves}

\begin{figure}[t]
\center{\includegraphics[width=\linewidth]{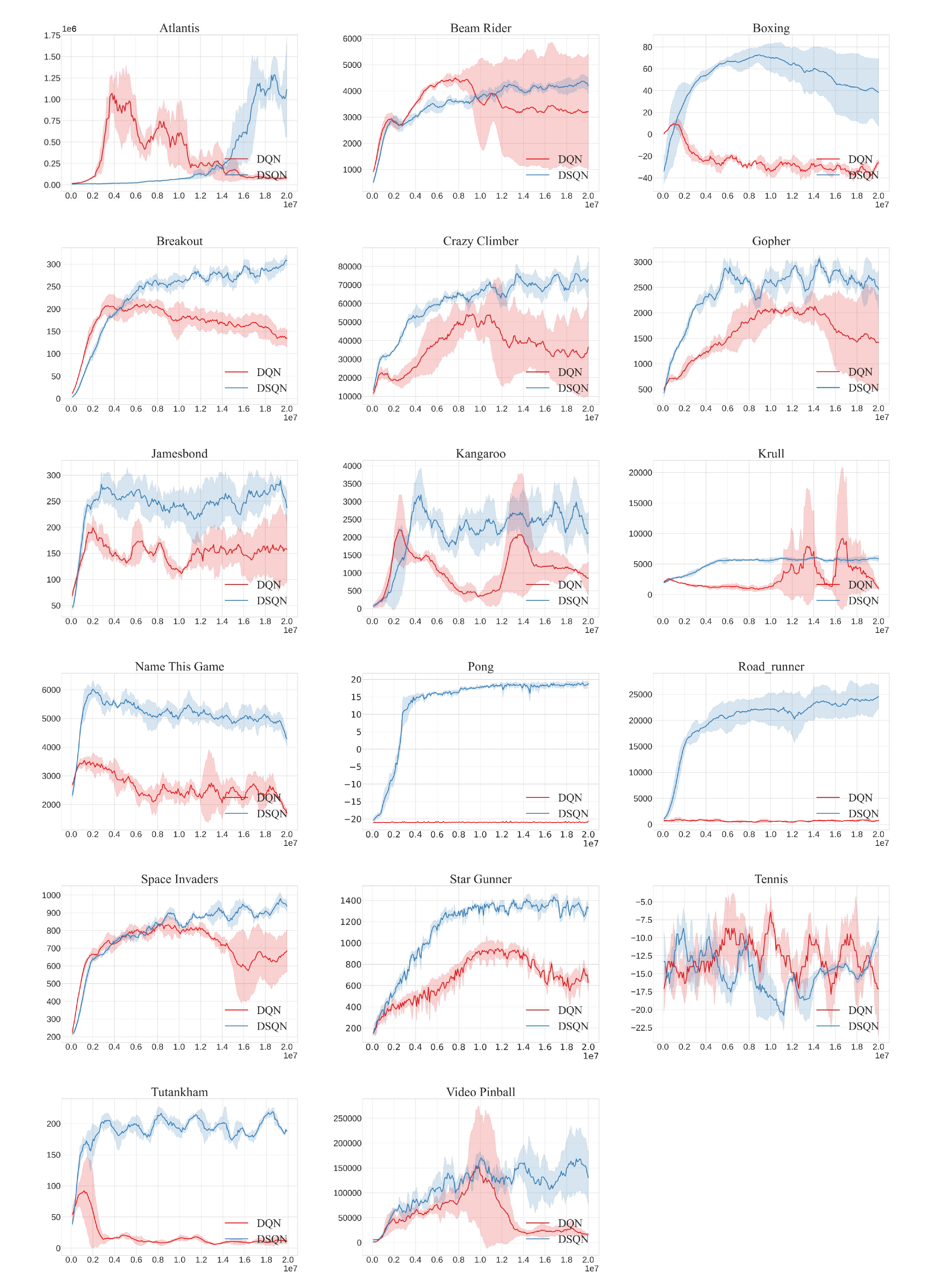}} 
\caption{ \label{fig:all_curves} Learning curves for DQNs and DSQNs on 17 Atari games. Every curve is smoothed with a moving average of 10 to improve readability. }
\end{figure}

In this section, we provide complete learning curves of DQNs and DSQNs on 17 Atari games. From complete learning curves (see Figure \ref{fig:all_curves}), it can be concluded that DSQNs can alleviate the decrease of scores caused by over-estimation in DQNs, which is one of the reasons why we only train for a total of 20 million frames.

\subsection{Detailed Tables of Scores}

To obtain summary statistics of each algorithm across games, we normalize the scores of each game as follows:
\begin{equation}
    score_{normalized}=\frac{score_{agent} - score_{random}}{score_{human} - score_{random}}.
\end{equation}

In Table \ref{tab:score}, we report the normalized scores of DQN, Converted SNN (ANN-SNN) and DSQN, where 0\% corresponds to a random agent and 100\% to the mean score of human experts. For DQN and ANN-SNN, we report our own re-runs. Since the performance of agents in different games is quite different, the median human normalized performance across all games is usually used to compare different algorithms. From the table, we obtain the median human normalized performance: 142.8\% (DQN), 96.2\% (ANN-SNN) and 193.5\% (DSQN), which demonstrates that our method outperforms DQN, using fewer time-steps.

\begin{table*}[t]
\caption{The normalized scores of Atari games obtained by DQN agents, Converted SNN (ANN-SNN) agents and DSQN agents. The simulation time $T$ is set to 500 (ANN-SNN) and 8 (DSQN).}
\label{tab:score}
\centering
\begin{tabular}{lrrrrr}
\toprule
Game & Random & Human & DQN & ANN-SNN & DSQN \\
\midrule
Atlantis & 12850.0 & 29028.0 & 17054.1\% & 9752.2\% & 15472.1\% \\
Beam Rider & 363.9 & 5775.0 & 101.2\% & 83.9\% & 91.7\% \\
Boxing & 0.1 & 4.3 & 502.4\% & 422.5\% & 1966.9\% \\
Breakout & 1.7 & 31.8 & 862.9\% & 671.9\% & 1217.1\% \\
Crazy Climber & 10781.0 & 35411.0 & 302.9\% & 255.7\% & 342.6\% \\
Gopher & 257.6 & 2321.0 & 142.8\% & 100.6\% & 192.7\% \\
Jamesbond & 29.0 & 406.7 & 129.6\% & 79.0\% & 116.6\% \\
Kangaroo & 52.0 & 3035.0 & 158.9\% & 133.5\% & 193.5\% \\
Krull & 1598.0 & 2395.0 & 4153.1\% & 3262.1\% & 676.7\% \\
Name This Game & 2292.0 & 4076.0 & 151.3\% & 125.0\% & 262.8\% \\
Pong & -20.7 & 9.3 & 0.6\% & 0.0\% & 134.0\% \\
Road Runner & 11.5 & 7845.0 & 33.0\% & 28.8\% & 353.8\% \\
Space Invaders & 148.0 & 1652.0 & 59.3\% & 47.8\% & 70.1\% \\
Star Gunner & 664.0 & 10250.0 & 4.9\% & 4.4\% & 10.4\% \\
Tennis & -23.8 & -0.1 & 96.2\% & 96.2\% & 96.2\% \\
Tutankham & 11.4 & 167.6 & 92.9\% & 78.0\% & 163.3\% \\
Video Pinball & 16257.0 & 17298.0 & 28286.5\% & -357.7\% & 37634.5\% \\
\midrule
\textbf{Median} &  &  & \textbf{142.8\%} & \textbf{96.2\%} & \textbf{193.5\%} \\
\bottomrule
\end{tabular}
\end{table*}

\subsection{Theoretical Energy Computation}

We calculate the theoretical energy consumption of convolutional layer and fully-connected layer used in the network with different activation functions and neuronal models. For fully-connected layers, we assume that the dimension of input and output is $m$ and $n$ respectively. For convolutional layers, we assume that the kernal size is $K \times K$, the number of input and output channels is $C_{in}$ and $C_{out}$ respectively, the length and width of the output feature map are $H_{out}$ and $W_{out}$ respectively, and the average spike probability of input is $fr_{in}$. $E_{AC}$ and $E_{MAC}$ denote the energy cost for AC and MAC operations, respectively.

\begin{table}[H]
\caption{The theoretical energy computation of different layers.}
\label{tab:energy}
\centering
\begin{tabular}{lr}
\toprule
Layer & Theoretical energy computation \\
\midrule
FC+ReLU/Identity    & $mnE_{MAC}$ \\
FC+LIF/LI           & $mnfr_{in}E_{AC}+nE_{MAC}$ \\
Conv+ReLU   & $K^2C_{in}C_{out}H_{out}W_{out}E_{MAC}$ \\
Conv+LIF    & $K^2C_{in}C_{out}H_{out}W_{out}fr_{in}E_{AC}+C_{out}H_{out}W_{out}E_{MAC}$ \\
\bottomrule
\end{tabular}
\end{table}

\end{document}